\documentclass[letterpaper, 10 pt, conference]{ieeeconf}  

\IEEEoverridecommandlockouts                              

\usepackage[table]{xcolor}
\usepackage{graphics} 
\usepackage{epsfig} 
\usepackage{mathptmx} 
\usepackage{times} 
\usepackage{amsmath} 
\usepackage{amssymb}  
\usepackage{xcolor}
\usepackage[colorlinks=true, linkcolor=blue, anchorcolor=blue, citecolor=blue, filecolor=blue, menucolor=blue, urlcolor=blue]{hyperref}
\usepackage{siunitx}
\usepackage{seqsplit}
\usepackage{subcaption}
\usepackage{multirow}
\usepackage{amsfonts}
\usepackage{bm}
\usepackage{placeins}

\DeclareMathOperator*{\argmin}{argmin}

\title{\LARGE \bf
Semantic Nearest Neighbor Fields for Monocular Edge Visual-Odometry
}

\author{Xiaolong Wu, Assia Benbihi, Antoine Richard, and C\'edric Pradalier
}

\begin{document}

\maketitle
\thispagestyle{empty}
\pagestyle{empty}

\begin{abstract}
  Recent advances in deep learning for edge detection and segmentation opens up
  a new path for semantic-edge-based ego-motion estimation.
  In this work, we propose a robust monocular visual odometry
  (VO) framework using category-aware semantic edges. It can reconstruct
  large-scale semantic maps in challenging outdoor environments. The core of
  our approach is a semantic nearest neighbor field that facilitates a robust
  data association of edges across frames using semantics. This significantly
  enlarges the convergence radius during tracking phases. The proposed edge
  registration method can be easily integrated into direct VO frameworks to
  estimate photometrically, geometrically, and semantically consistent camera
  motions. Different types of edges are evaluated
  and extensive experiments demonstrate that our proposed system outperforms
  state-of-art indirect, direct, and semantic monocular VO systems. 
\end{abstract}

\section{INTRODUCTION}
\label{sec:introduction}

In recent decades, monocular Visual Odometry (VO) and Simultaneous Localization
and Mapping (SLAM) systems have shown their full potential to assist various
robotic applications in outdoor operations - from autonomous driving in urban
scenes to environmental monitoring in natural environments. Among these
algorithms, indirect methods \cite{klein2007parallel} \cite{strasdat2010scale}
\cite{mur2017orb} are the \textit{de facto} standards since visual
features provide considerable robustness to both photometric noise and
geometric distortion in images. Recent works have shown that direct
methods \cite{newcombe2011dtam} \cite{pizzoli2014remode} \cite{engel2014lsd}
\cite{engel2018direct} could provide more accurate and robust motion
estimation. 
However, they present a much smaller convergence basin compared with indirect 
methods because of loose data association. 

Edge-based ego-motion estimation has also gained significant attention for its
robustness against illumination changes, motion blur, and occlusion. Edge VO and
SLAM can be seen as a crossover of indirect and direct principles.
Specifically, edges are binary features extracted from raw images, but edge
registration is performed using iterative-closest-point (ICP) based direct
alignment. Motion estimation using edges is particularly attractive for outdoor
applications for its illumination and convergence robustness. Still,
standard edge detection methods (e.g. Canny) used in edge VO and SLAM provide edges
with poor repeatability in outdoor settings, which hinders their usage.
Recent developments of learning-based edge detection opens up a new path 
to improve the stability of edge detection for outdoor edge VO and SLAM.

In addition to feature and edge learning, semantic information can also boost motion
estimation. \cite{lianos2018vso} integrates a semantic reprojection error 
into state-of-art indirect and direct VO systems to improve the
tracking robustness and accuracy. However, it may be not suitable for
edge-based motion estimation because of ambiguous semantic labels of edges
around semantic boundaries. 

\begin{figure}[t]
\centering
\includegraphics[width=\linewidth]{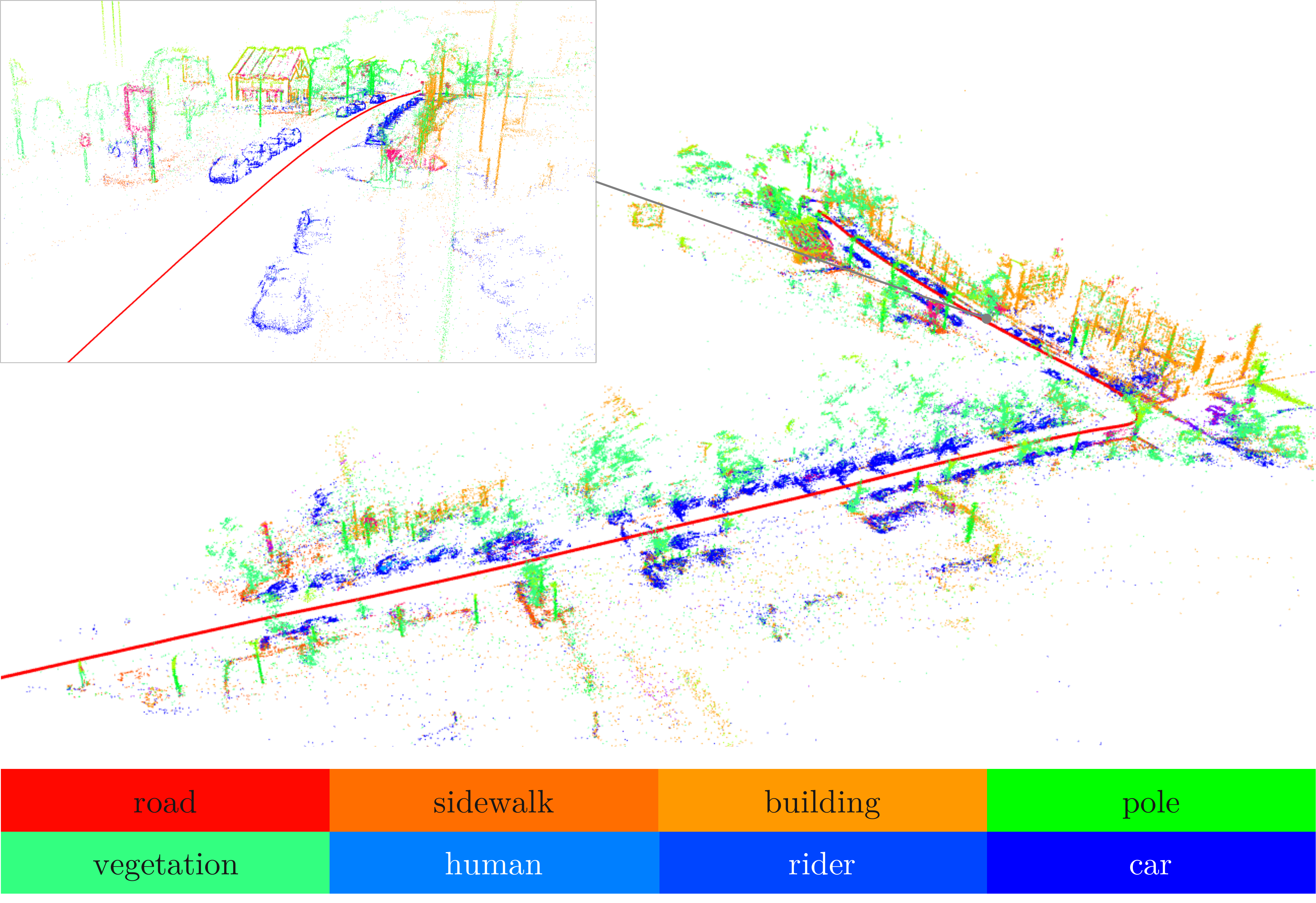}
  \caption{3D semantic map with our semantic edge VO
  system. Semantic edges are learned with CaseNet~\cite{yu2017casenet}.}
\label{fig:semanticmap}
\end{figure}
 
In this work, we present an edge-based VO framework that can track and
reconstruct semantic edges across frames for outdoor robotic applications. We
show that our proposed semantic nearest neighbor fields (SNNFs) offer several
advantages over existing edge-based VO algorithms. The learned semantic edges
further improve the accuracy and robustness, especially in outdoor environments
in Fig.~\ref{fig:semanticmap}. We investigate the influence of several edge
detection methods on motion estimation. We conduct extensive evaluation on KITTI,
a public autonomous driving dataset, and measure tracking accuracy,
robustness and runtime performances. Experimental results show that our
proposed system outperforms state-of-art monocular direct, indirect, and
semantic VO systems in an outdoor setting. Our main contributions are:

\begin{itemize}
  \item A novel SNNFs strategy using semantics to improve the efficiency and
    robustness of ICP-based edge registration.
  \item A monocular semantic edge VO system that integrates image gradient,
    edges and semantic information to compute photometrically,
    geometrically, and semantically consistent motion. It is
    capable of reconstructing large-scale semantic maps. 
  \item A evaluation study of several edge detection methods for outdoor ego-motion
    estimation.  To our knowledge, this is the first paper evaluating 
    edge-based VO system in an outdoor environment.  
\end{itemize}  

\section{RELATED WORK}
\label{sec:relatedwork}

VO and SLAM are two main solutions for camera tracking and environmental
mapping, and can be divided into indirect or direct methods.
Indirect methods \cite{klein2007parallel} \cite{mur2017orb} minimize the
reprojection error between features in two consecutive images. In contrast to
feature-based approaches, direct methods \cite{engel2013semi}
\cite{engel2014lsd} \cite{engel2018direct} minimize the photometric error
between all the matching pixels in two successive images. The main drawback 
is that it presents a much smaller convergence basin
than indirect methods because of its loose data association. 

Edge-based ego-motion estimation relies on ICP-based optimization.
\cite{kneip2015sdicp} proposes an efficient 2D-3D edge registration framework
for real-time motion estimation using distance transform (DT)
\cite{felzenszwalb2012distance}. However their objective function is neither
differentiable nor negative. \cite{kuse2016robust} relieves the first issue
with a sub-gradient method.  \cite{kim2018edge} solves the second and enables
Gauss-Newton like optimization with a signed residual based on 2-D edge
divergence minimization. \cite{zhou2017semi} tackles both problems concurrently
and substitutes DT with approximate nearest neighbor fields (ANNFs). The
subsequent work \cite{zhou2019canny} extends ANNFs into oriented nearest
neighbor fields (ONNFs). It provides a finer data association strategy to
reject outliers and enables more robust edge registration.  Besides pure
edge-based methods, \cite{wang2016edge} and \cite{li2016fast} combine
intensity-based photometric and edge-based geometric errors to improve tracking
robustness and accuracy. However, performance may degrade with illumination
changes contrary to semantic information that is more invariant.

With the recent advances in deep learning, semantic information has become
relevant for motion estimation.  \cite{yu2018ds} \cite{mahe2018semantic} use
semantics to detect moving objects and alleviate their pixels weights in the
objective function.  Similarly, \cite{kaneko2018mask} masks the sky pixels.
\cite{lianos2018vso} integrates a semantic reprojection error into existing
point-based indirect and direct motion estimation systems. The overall
advantage of integrating semantics is to
boost the tracking and mapping robustness and accuracy. We pursue these efforts
and fuse the advantages from both semantic-SLAM and learned-edges-SLAM to make
VO even more robust.

Previous edge-based SLAM methods rely on the standard Canny edge detector
\cite{canny1987computational}, but \cite{schenk2017robust} shows that using
machine-learned edges \cite{dollar2015fast} \cite{xie2015holistically} gives
significant improvements compared to
standard edges in indoor settings. Structured edges (SE) \cite{dollar2015fast}
generalizes random forests to general structured output spaces to learn edges
on patch images. Despite its fast computation time, it relies on hand-crafted
features and requires to map the ground-truth edges to low dimensional
representation for the training to be scalable. These drawbacks are alleviated
with the deep-learning based HED \cite{xie2015holistically}: it processes raw
images and outputs an edge probability map over the input in the form of a
binary map. In this work, we extend edge-based SLAM and enhance them with
semantic information to make them more robust. To do so, we use learned-edges
together with their semantic labels as computed in \cite{yu2017casenet}
\cite{yu2018simultaneous}.
CaseNet \cite{yu2017casenet} extends HED \cite{xie2015holistically} to multi-label edges
and outputs a distinct probability edge map for each semantic class. 
SEAL \cite{yu2018simultaneous} continues the efforts of \cite{yu2017casenet}
and tackles the challenge of edge-alignment during the CNN training. Label
noise in the human-annotated ground-truth edges leads CaseNet to learn thick
edges which hides relevant edges details. However, none of these works investigate the
application of semantic edges in VO. In contrast, we provide a method to
integrate any semantic edge into VO and run extensive experiments to compare
the influence of edge and semantic learning on the VO performances.

\section{SEMANTIC EDGE VISUAL ODOMETRY}
\label{sec:edgevo}

We first give a brief introduction on edge-based motion
estimation and discuss the main limitation of existing work in Sec.
\ref{sec:problemformulation}. We then introduce our proposed SNNFs and the
corresponding edge registration algorithm in Sec.~\ref{sec:snnf}. The
implementation details are provided in Sec.~\ref{sec:vo}.

\begin{figure*}[t]
\centering
\includegraphics[width=\linewidth]{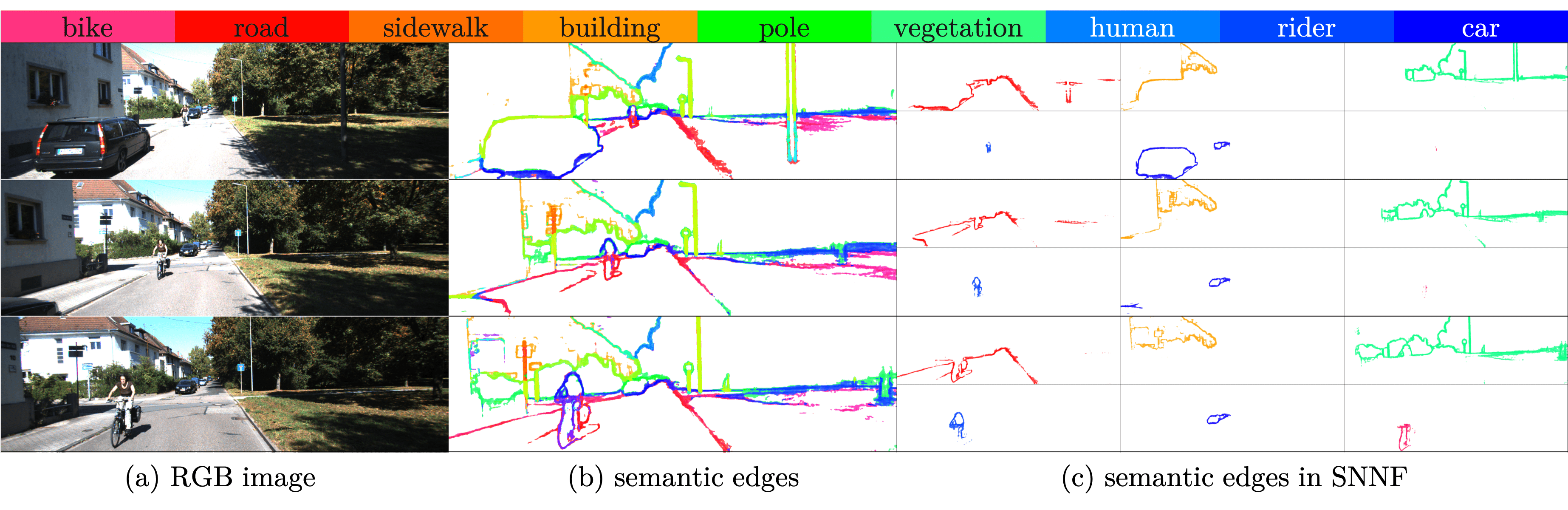}
  \caption{Example sequential RGB images, fused semantic edge maps, and their SNNFs.  }
\label{fig:snnf}
\end{figure*}

\subsection{Problem Formulation} 
\label{sec:problemformulation}

Let us consider a reference frame, a gray-scale reference image $I_{r}: \Omega
\rightarrow \mathbb{R}$ and an inverse depth map  $D_{r}: \Omega \rightarrow
\mathbb{R}^{+}$, where $\Omega \subset \mathbb{R}^2$ is the image domain. A 3D
scene point $\mathbf{P} = (x, y, z)^{T}$ is parameterized by its inverse depth
$d = z^{-1}$ in the reference frame instead of the conventional 3 unknowns.
Thus, each pixel in the reference frame $\mathbf{p} = (u,v)^T \in \Omega$ can
be back-projected into 3D world using the back-projection function
$\pi^{-1}(\cdot)$ as:
\begin{equation} \label{eq:eq1}
\mathbf{P} =  \pi^{-1}(\mathbf{p},d) = \mathbf{K}^{-1}\mathbf{\bar{p}}/d
\end{equation}
where $\bar{\mathbf{p}} = (\mathbf{p}^T,1)^T $ is the homogeneous coordinate of
pixel coordinate and $\mathbf{K}$ is the pre-calibrated camera intrinsic
matrix. Inversely, the 3D projective warp function $\pi(\cdot)$ can be
expressed as:
\begin{equation} \label{eq:eq2}
\mathbf{p} =  \pi(\mathbf{P}) = \mathbf{\bar{K}} 
\begin{bmatrix}
    x/z  \\  y/z  \\
\end{bmatrix}^{T}
\end{equation} 
where $\mathbf{\bar{K}}$ is the first two rows of intrinsic matrix $\mathbf{K}$.

Given arbitrary edge detector $\mathit{E}(\cdot)$, the group of edge pixels
$\mathcal{E}_r$ extracted from the reference image can be expressed as:
\begin{equation} \label{eq:eq4}
\mathcal{E}_r =  \left \{ \mathbf{p}_r \right \} = \mathit{E}(I_r) 
\end{equation} 
The detected edge pixels $\mathcal{E}_{r}$ are subsequently projected to
current frame $k$. The transformed edge pixels $\mathcal{E}_{kr}$ can be
computed as:
\begin{equation} \label{eq:eq5}
\mathcal{E}_{kr} = \left \{ \mathbf{p}_{kr} \right \}= \left \{ \pi ( \mathbf{R}_{kr}\pi^{-1}(\mathbf{p}_r,D_{r}(\mathbf{p}_r) )+\mathbf{t}_{kr} ) \right \}
\end{equation} 
where $ \mathbf{R}_{kr} \in SO(3)$ and $\mathbf{t}_{kr} \in \mathbb{R}^3$ are
the 3D rigid body rotation and translation from reference frame to current
frame $k$, respectively. 

Let us define a function to find the nearest neighbor of the projected edge
pixel $\mathbf{p}_{kr}$ in current frame $k$ using the Euclidean distance
metric as:
\begin{equation} \label{eq:eq6}
n(\mathbf{p}_{kr} ) = \argmin_{ \mathbf{p}_{k} \in \mathcal{E}_k } \Vert \mathbf{p}_{k} - \mathbf{p}_{kr} \Vert
\end{equation}
As the result, the total energy $E_{kr}^{\mathcal{E}}$ sums up all edge
distance errors from the reference frame to the current frame $k$ expressed as:
\begin{equation} \label{eq:eq7}
E_{kr}^{\mathcal{E}} :=  \sum_{\mathbf{p}_{r} \in \mathcal{E}_r } w_{\mathbf{p}_r}^{\mathcal{E}} \Vert \mathbf{p}_{kr} - n(\mathbf{p}_{kr}) \Vert_{\gamma}
\end{equation}
where $w_{\mathbf{p}_r}$ is the weight assigned for each edge pixel in the
reference frame and $\Vert \cdot \Vert_{\gamma}$ is the Huber norm.  

To find the optimal camera transformation $\mathbf{R}$ and $\mathbf{t}$ using
Eqn. \ref{eq:eq6}, we use a 2D-3D ICP-based optimization \cite{kneip2015sdicp}
that alternates between finding approximate nearest neighbors and
register the putative correspondences using an iteratively reweighted
Gauss-Newton algorithm.  Following the theory of optimization under unitary
constraints \cite{manton2002optimization}, we optimize the energy function
defined in Eqn.  \ref{eq:eq6} on Lie-manifolds for better convergence. 

\subsection{Semantic Nearest Neighbor Fields} 
\label{sec:snnf}

ANNFs-base edge VO is extensively evaluated in \cite{zhou2017semi} on an indoor
dataset, and presents impressive performance. However, when it comes to outdoor
environments, the results of both ANNFs and ONNFs deteriorate
because of unrepeatable edges and large camera motions. This comes from
the weak data association strategy of edge-based VO algorithms: it makes
them sensitive to outliers and motion initialization. To solve this issue, we
incorporate semantic information into existing edge VO frameworks and facilitate
a robust data association of edges across frames.

The key idea behind SNNFs is to classify the extracted edges as shown in
Fig.\ref{fig:snnf}. Then each group of edges can only be registered to their
counterparts in subsequent frames. Specifically, for a dataset with $C$
classes, we generate $C$ edge-class probability maps. An edge pixel can belong
to one or more class: for example, an edge pixel lying between a car and the
road has both labels. Then, we compute distance fields with region growing
algorithm on the seeded region of each map. Upon
registration, each edge pixels is only registered with edges with the same
semantic label instead of spatially nearest ones. This way, we avoid ambiguous
associations and enlarge the convergence basin during registration: 
SNNFs constrains edge registration with semantic consistency so the
distance to the region of attraction that is semantically and geometrically
consistent is much larger than for ANNFs. Note that our proposed SNNFs implements
a 'soft' data association strategy: each pixel can be classified
into one or multiple semantic classes. This allows our method to be robust to
edge classification errors.

SNNFs is, in essence, an extension of the ANNFs, so it can be seamlessly
integrated to the optimization formulation in Eqn. \ref{eq:eq7}. The new energy to
optimize is: 
\begin{equation} \label{eq:eq10}
E_{kr}^{\mathcal{E}} :=  \sum_{i = 1}^{C}\sum_{\mathbf{p}_{r} \in {\mathcal{C}_r}^i } w_{\mathbf{p}_r}^{\mathcal{E}} \Vert \mathbf{p}_{kr} - n_{{\mathcal{C}_r}^i} (\mathbf{p}_{kr}) \Vert_{\gamma}
\end{equation}
where ${\mathcal{C}_r}^i$ denotes the subset of edge pixels belonging to
$i^{th}$ semantic class, $n_{{\mathcal{C}_r}^i}(\cdot)$ is the function returns
the nearest neighbor from the $i^{th}$ semantic edge group in frame $k$. The 
relationship between $\lbrace {\mathcal{C}_r}^i \rbrace$ and
$\mathcal{E}_r$ can be expressed as $ {\mathcal{C}_r}^1\cup \cdots \cup
{\mathcal{C}_r}^C = \mathcal{E}_r $.

Following \cite{zhou2017semi}, we implement a point-to-tangent residual i.e. 
we project the original pixel-wise residual onto its local gradient
direction to obtain additional robustness against outliers. It should be noted
that this formulation makes the underlying assumption that the camera motion is
free of large inter-frame rotations. In reality, this assumption is valid for
the autonomous driving applications considered in this paper. 

\subsection{Implementation Detail }
\label{sec:vo} 

We integrate the semantic edge constraints (Eqn. \ref{eq:eq10}) into both
tracking and mapping. In the tracking phase, we put more weights on the edge
residuals to enforce a better convergence basin. In the mapping phase, we set
smaller weights for edge terms and use a depth regularization component to
penalize large inverse depth updates: the inverse depth of some edge
pixels may be unobservable as the epipolar line are perpendicular to edge
normals. 

We use image gradient magnitude instead of intensity as it is more robust to
illumination changes and preserves the high-frequency photometric information.
Edge-based VO is robust to light variations but relying on image intensity 
may jeopardize this property. This requires only a slight code modification.

As detailed in Sec. \ref{sec:evaledges}, we implement a
flexible point selection strategy to boost the tracking performances in
vegetation dominated environments. Specifically, our proposed semantic edge VO
algorithm not only selects edge pixels with semantic labels but also randomly
select unlabeled pixels as support pixels to boost the robustness of motion
estimation. When the number of edge pixels is large enough, the pixel selector
merely samples the minimum number of supportive pixels. However, as the firm
edges get less or only occupied in small areas, we select more supportive
points to keep well-distributed pixels into the optimization layer. We use the
same sampling strategy described in \cite{engel2018direct} to choose the
supportive pixels. Note that the supportive pixels only have photometric
constraint while edge pixels hold both edge and photometric constraints. Since
the supportive points don't have any labels, we don't push them into the
semantic maps. 

\section{EXPERIMENTS}
\label{sec:experiment} 

We first describe the experimental setup and the dataset. Qualitative results
on semantic mapping results are presented in Sec. \ref{sec:semanticmapping}.
Sec. \ref{sec:parameterstudy} evaluates the influence of several edge detection
and registration algorithms on tracking accuracy and robustness.
Our proposed semantic edge VO is compared to the state-of-art edge and
semantic VO systems on KITTI in Sec. \ref{sec:evalvo}. 
Finally, the runtime performance is discussed in Sec. \ref{sec:runtime}.

\subsection{Dataset and Metric} 
\label{sec:dataset}

We test our method on sequences 00-10 of the KITTI odometry dataset
\cite{geiger2012we}. We use the rectified color image of the left camera only with
intrinsic calibration and ground truth camera poses. 

\begin{table}[h]
\centering
\caption{ KITTI Dataset Description}
\begin{tabular}{ccc}
\hline
Scene   & Sequence No.        & Description                              \\ \hline
city    & 00, 05, 06, 07      & buildings, cars with few vegetation  \\
village & 02, 03, 04,  08, 09 & vegetation with few buildings, cars \\
highway & 01                  & roads, cars, and signs                   \\ \hline
\end{tabular}
\label{table:dataset}
\end{table}

For fair comparison, we do not use loop-closure and all methods use the ground
truth poses to recover the scale of motion every 200 frames. During evaluation,
we find that the pose estimation for the first frames are unstable and vary
for each method. So we discard the first ten pose estimates for all
experiments. We choose the absolute trajectory error (ATE) as our metric: it
measures the absolute difference between camera positions of two trajectories.

\subsection{Semantic Mapping} 
\label{sec:semanticmapping}

\begin{figure*}[htb]
  \centering
  \includegraphics[width=\linewidth]{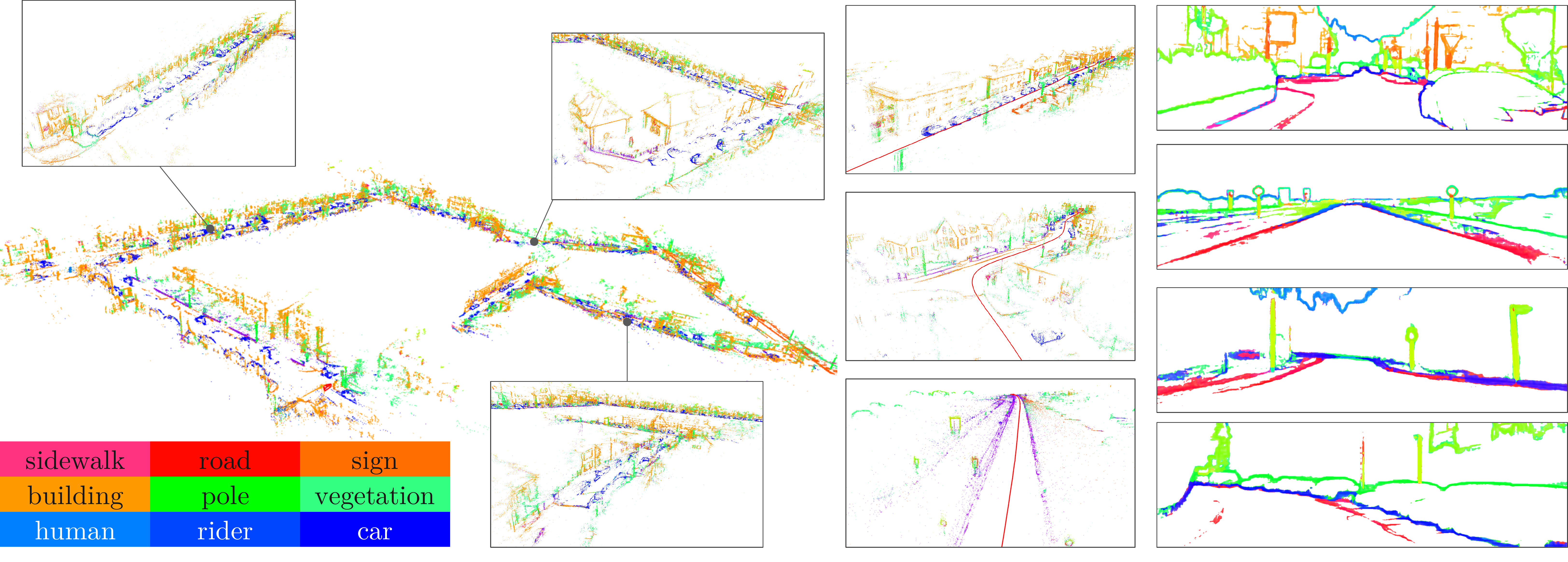}
  \caption{ Reconstructed semantic edge maps for KITTI. Left: semantic edge 
  maps recovered from city, village, and highway sequences. Right:
  semantic edge images generated using CaseNet \cite{yu2017casenet}.}
  \label{fig:reconstruction}
\end{figure*}

Fig. \ref{fig:reconstruction} shows large-scale semantic edge maps with pixel
resolution generated with our approach on several environments.  
It should be noted that our proposed
monocular VO system can only recover locally consistent 3D semantic edge maps
rather than global ones since our algorithm also suffers from scale drift like
all monocular methods.  

\begin{figure}[htb]
  \centering
\includegraphics[width=\linewidth]{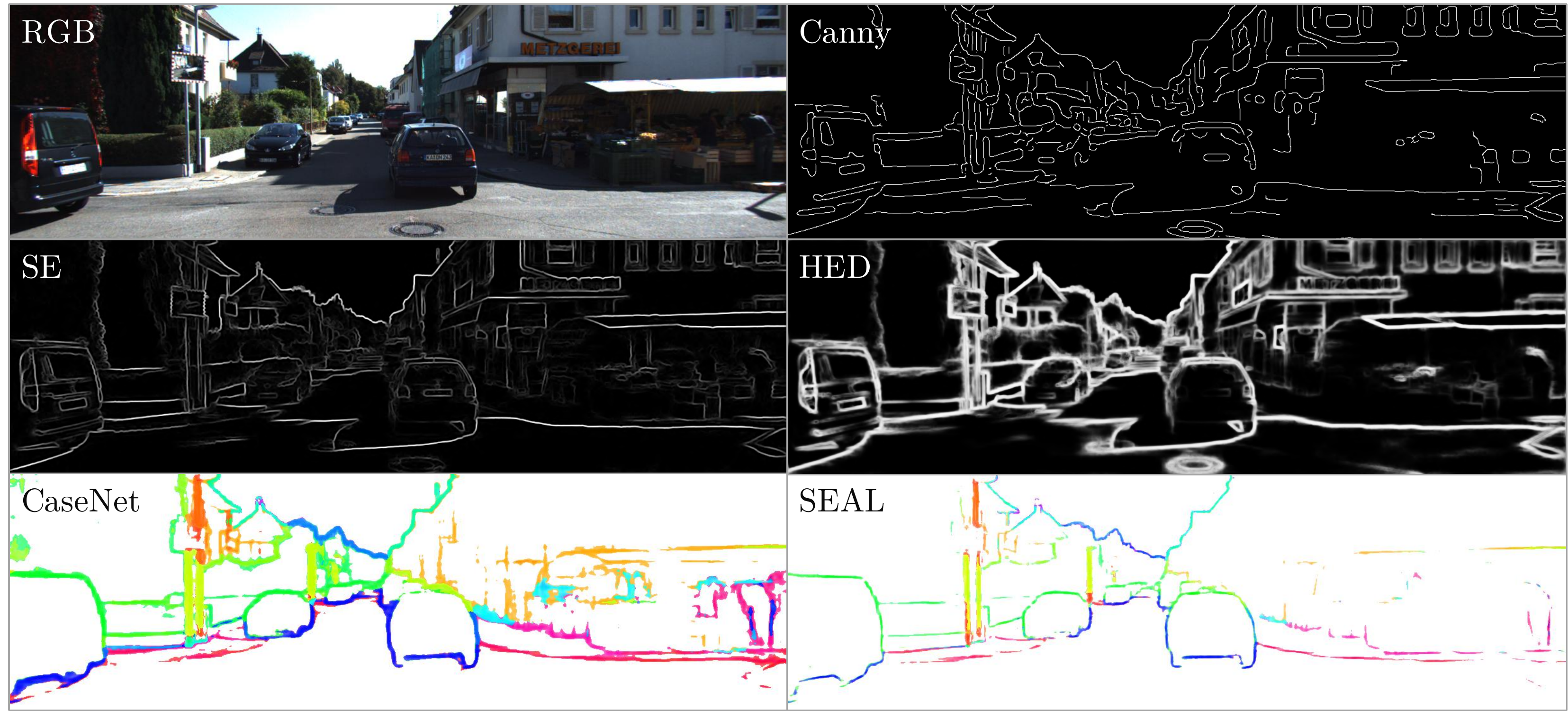}
  \caption{Visualization of the evaluated edge methods.}
\label{fig:edges}
\end{figure}

\subsection{Edge-learning evaluation} 
\label{sec:parameterstudy}
We investigate the influence of the edge learning and registration methods
on the robustness and accuracy of SNNFs (Fig. \ref{fig:edges}). 

\subsubsection{Edge Repeatability} 
\label{sec:repeatabilityanalysis}

\begin{figure*}[htb]
  \centering
  \includegraphics[width=\linewidth]{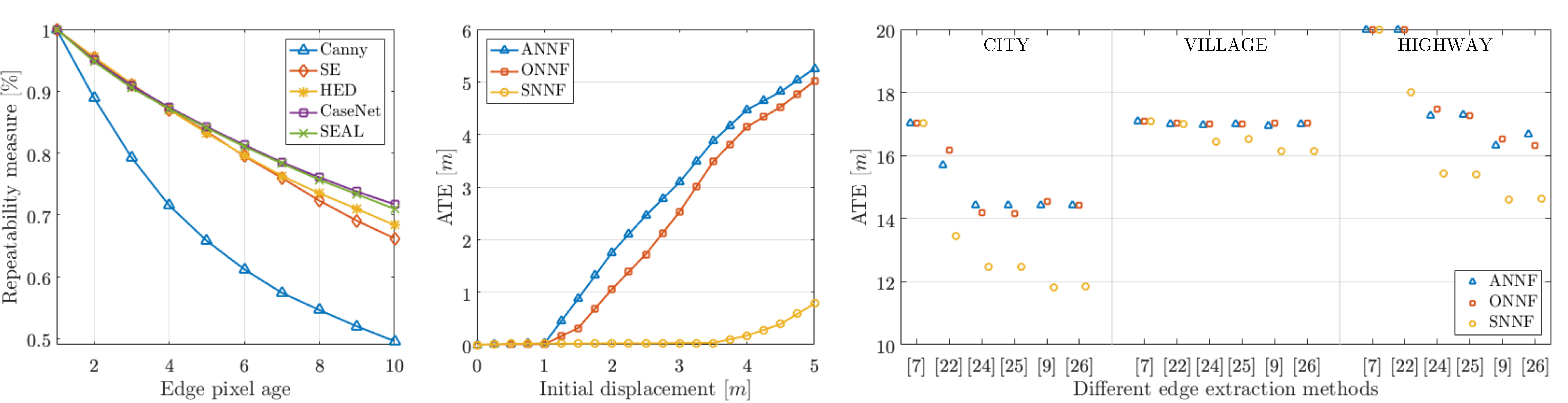}
  \caption{ Left: Repeatability analysis on vKITTI. Middle: convergence basins of ANNFs, ONNFs, and our proposed SNNFs edge
  registration on vKITTI.
  Right: tracking errors for different
  registration methods and different edge generation methods (city, village,  highway scenes). We compare conventional edge detector
  (Canny \cite{canny1987computational}), learned edges (SE
  \cite{dollar2015fast}, HED \cite{xie2015holistically}), and semantic edges
  (CaseNet \cite{yu2017casenet}, SEAL \cite{yu2018simultaneous}). DSO \cite{engel2018direct} serves as a baseline VO.
  }
  \label{fig:registration}
\end{figure*}

For outdoor VO applications, the choice of edge detector $\mathit{E}(\cdot)$ in
Eqn. \ref{eq:eq4} is still an open question. \cite{schenk2017robust} observes
that the performance of edge-based VO highly depends on the
\textit{repeatability} of the edges: it computes the number of re-detected
edge-pixels over the number of edge-pixels that should reappear. For fair
comparison, we randomly choose 9000 edge pixels for each edge detectors. We run 
evaluation on vKITTI \cite{gaidon2016virtual} since the ground truth depth is
available.  Fig. \ref{fig:registration} shows the repeatability analysis: 
learned edges significantly outperform the conventional Canny detector.  

\subsubsection{Edge Registration Evaluation} 
\label{sec:evalregistration}

One of the main advantage of our method is to improve tracking robustness and
accuracy by integrating semantic information in the optimization.  Experiments
show that the convergence basin of our method is indeed larger than
state-of-art ANNFs- and ONNFs-based methods \cite{zhou2019canny}. 

Fig. \ref{fig:registration} shows ATE for ANNFs, ONNFs, and SNNFs based edge
registration approaches as a function of initial displacement using vKITTI
dataset. We use the ground-truth depth maps to rule out the bias of depth
reconstruction.  We first use the ground truth poses as initial camera poses
and control the initial displacements in a range of $[0,5]$m.  The ATE is
obtained by averaging all inter-frame tracking errors to compute the ATE.  We
observe that the convergence basin of our proposed SNNFs-based method is about
two times larger than that of ONNFs and ANNFs. Note that the convergence test
is not implementing any multi-level optimization techniques to improve the
convergence basin. But our proposed VO system implements a pyramid-based
tracking strategy to boost the tracking robustness further.

\subsubsection{Edge-Learning Evaluation} 
\label{sec:evaledges}
We evaluate the influence of the edge-learning method on the VO performances.
Fig~\ref{fig:registration} (bottom-right) shows the ATE on KITTI for ANNFs,
ONNFs, and SNNFs using conventional edges, fused-semantic-edges and 
learned-semantic-edges.

We distinguish two ways to generate semantic edges: one is end-to-end learned
semantic edges such as CaseNet and SEAL trained on Cityscapes
\cite{cordts2016cityscapes}. The second one is the fusion of the learned-edges,
HED or SE trained on BSDS \cite{arbelaez2011contour}, and learned-semantics trained
separately from state-of-the-art DeepLabV3 \cite{deeplabv3plus2018}.  We use
the Xception-65 \cite{chollet2017xception} model pretrained on Cityscapes and
finetuned on
KITTI~\footnote{\url{https://github.com/hiwad-aziz/kitti\_deeplab}}. We assume
that edge and segmentation probability are independent and multiply them to
compute the fused semantic-edge probabilities.
We observe that CaseNet and SEAL generalize well from Cityscapes to KITTI
without further finetuning contrary to dense segmentation. This is a strong
advantage of end-to-end methods over fusion ones.
It
should be noted that the edge VO algorithms implemented in this evaluation
merely use edge pixels rather than the more flexible ones described in Sec.
\ref{sec:vo}, so that we can observe the pros and cons of integrating edge and
semantic constraints in outdoor VO applications.  

\begin{figure*}[htb]
  \centering
  \includegraphics[width=\linewidth]{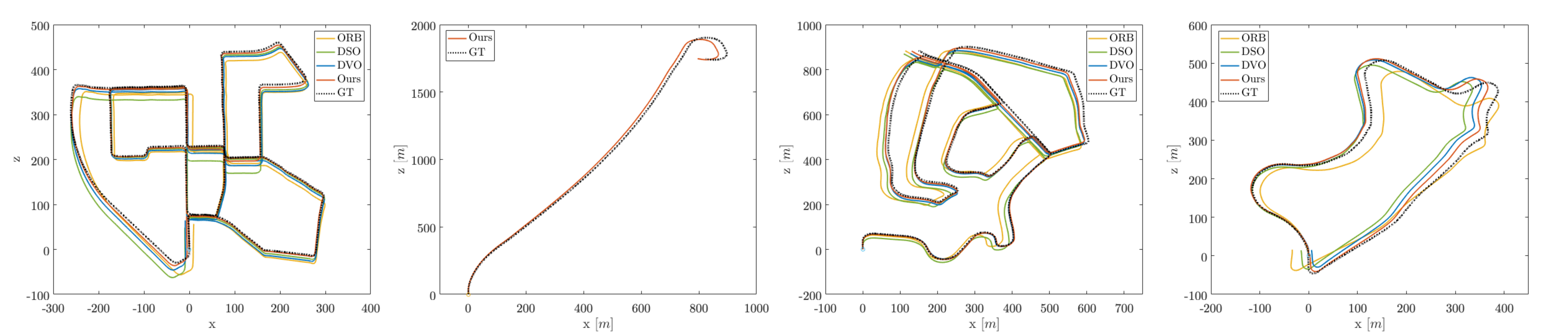}
  \caption{Trajectories from our method, indirect ORBSLAM2, direct DSO, and semantic VSO systems on KITTI.
  Left to Right: KITTI-seq00, 01, 02, and 09. Note that
  seq01 merely shows our trajectory and the ground
  truth bcause other methods cannot generate the whole trajectory. }
  \label{fig:trajectory}
\end{figure*}

\begin{table*}[htb]
\centering
\caption{ Tracking Error Comparison}
\begin{tabular}{rcccccccccc}
\hline
\multicolumn{1}{c}{\multirow{2}{*}{KITTI}} & \multicolumn{4}{c}{city} & \multicolumn{5}{c}{village}      & highway \\
\multicolumn{1}{c}{}               &     00            & 05              & 06               & 07            & 02               & 03             & 04            & 08              & 09              & 01      \\ \hline
ORBSLAM2                              &  16.14          & 15.96         & 13.35          & 10.63        & 15.58         & \textbf{3.44} & 3.05         &  15.43      & 12.88          &  36.32      \\
DSO                                        & 16.83           & 13.64         & 16.83          & 9.55          & 17.08         & 3.71 & \textbf{3.01} & 18.31     & 13.05          & -       \\
SVO                                        & 15.31           & 10.08         & 14.10          & 8.39          & 14.57         & 3.76          & 3.09          & 15.29      & 13.12          & -       \\
Proposed                                 & \textbf{11.82}  & \textbf{8.39}  & \textbf{10.92} & \textbf{6.11} & \textbf{14.15} & 3.72          & 3.03      & \textbf{15.07} & \textbf{12.63} & \textbf{14.59}   \\ \hline
\end{tabular}                    
\label{table:eval}
\end{table*}

This evaluation allows us to draw five significant observations: (1)
Introducing edge constraints into VO system is advantageous for operation in
city and highway scenes, where there are many easy-to-track edges.  However, it
shows up little improvements or marginally worse performance for village
datasets due to the poor repeatability and the few edges in
vegetation-dominated images. (2) The VO systems using learned edges show a
slightly better tracking accuracy compared with the one uses conventional edge
detectors. This can be explained with the better repeatability of learned edges
in outdoor environments. (3) Semantic information benefits camera tracking more
in city and highway scenes, where there are more semantic elements than in non
urban areas.  The performances are barely improved for the village images due
to the edge sampling strategy. (4) End-to-end semantic edges have higher
tracking accuracy than the fused ones which suggests that the end-to-end
learning is more stable. (5) Introducing edge constraints cannot significantly
improves the overall tracking precision in village image sequences, which
indicates that merely relying on the edges in vegetation dominated environments
could potentially jeopardize the overall tracking robustness. 

These evaluations show that edge and semantic constraints significantly improve
the tracking performance for environments with semantic elements and edges such
as urban areas. However, it shows poor advantages in scenes with weak edges. In
comparison, DSO implements a flexible sampling strategy to achieve similar
tracking accuracy without incorporating edge and semantic constraints. We can
find that the main limitation of pure edge VO is the deficiency of
well-distributed edge pixels all over the image area. This justifies our
flexible sampling strategy to introduce pixels without labels to stabilize the
motion estimation in vegetation dominated environment in Sec. \ref{sec:vo}.

\subsection{Evaluation using Different Visual Odometry Methods} 
\label{sec:evalvo}

We assess the robustness and accuracy of the whole system
on KITTI. We compare to mono-ORBSLAM2 \cite{mur2017orb},
DSO \cite{engel2018direct} and VSO \cite{lianos2018vso} as state-of-the-art
monocular indirect, direct, and semantic VO algorithms for comparison. Since
there is no released code for VSO, we implement it by
introducing the semantic constraint energy into DSO for both tracking
and mapping. For fair comparison, we choose 4000 active points for
all approaches.
 
Table. \ref{table:eval} shows the quantitative results and  Fig.
\ref{fig:trajectory} shows trajectory comparisons on KITTI 00, 01, 02, and 09.
Our methods outperforms the state-of-art in terms of tracking accuracy.
Significant improvement can be observed on the highway trajectories, where only
our proposed method could recover the full trajectory for such environment.
Similar improvement are obtained on urban sequences and marginally better or
equivalent performance for village sequences. 

\subsection{Runtime} 
\label{sec:runtime}
Runtime depends on the semantic edge pixels and the image size. In our experiments, we keep the original KITTI and vKITTI image sizes (around (1242,375)$px$), and use 3000 edge pixels and 1000 supportive pixels to balance accuracy and robustness. The processing time for both tracking and mapping is $1.34\times$ longer on average than DSO using 4000 active pixels. The semantic edge generations runs at 0.7$s/image$ on an NVIDIA GTX1080Ti which makes our approach quasi-online.

\section{CONCLUSION}
\label{sec:conclusion}

In this work, we present a monocular semantic edge VO framework capable of
reconstructing 3D semantic edge maps in unstructured outdoor environments.  Our
proposed semantic nearest neighbor fields (SNNFs) offers several advantages
over existing edge VO algorithms by using deep-learned semantic as a robust
data association strategy. We analyse the influence of edge-learning and
alignment methods on edge-based motion estimation and overcome the
primary limitations of edge VO for outdoor application. An extensive evaluation
of the accuracy and the robustness is conducted on KITTI. Results show that our
method outperforms the state-of-art monocular direct, indirect, and
semantic VO systems. 

\addtolength{\textheight}{-10cm}   
                                  
\bibliographystyle{IEEEtran}
\bibliography{IEEEabrv}


\end{document}